\pdfoutput=1
\documentclass[conference]{IEEEtran}

\usepackage{cite}
\usepackage{algorithmic}
\usepackage{graphicx}
\usepackage{textcomp}
\usepackage{xcolor}

\usepackage{url}
\usepackage{dirtytalk}
\usepackage{booktabs}
\usepackage{dblfloatfix}
\usepackage[compact]{titlesec}

\def\BibTeX{{\rm B\kern-.05em{\sc i\kern-.025em b}\kern-.08em
    T\kern-.1667em\lower.7ex\hbox{E}\kern-.125emX}}
\newcommand{\name}{T\textsuperscript{3} }
\begin{document}

\title{T\textsuperscript{3}: Domain-Agnostic Neural Time-series Narration}

\author{\IEEEauthorblockN{Mandar Sharma}
\IEEEauthorblockA{\textit{Computer Science} \\
\textit{Virginia Tech}\\
mandarsharma@vt.edu}
\and
\IEEEauthorblockN{John S. Brownstein}
\IEEEauthorblockA{\textit{Boston Children’s Hospital} \\
\textit{Harvard Medical School}\\
john.brownstein@childrens.harvard.edu}
\and
\IEEEauthorblockN{Naren Ramakrishnan}
\IEEEauthorblockA{\textit{Computer Science} \\
\textit{Virginia Tech}\\
naren@cs.vt.edu}
}

\maketitle
\thispagestyle{plain}
\pagestyle{plain}

\begin{abstract}
The task of generating rich and fluent narratives that aptly describe the characteristics, trends, and anomalies of time-series data is invaluable to the sciences (geology, meteorology, epidemiology) or finance (trades, stocks, or sales and inventory). The efforts for time-series narration hitherto are domain-specific and use predefined templates that offer consistency but lead to mechanical narratives. We present \name (Time-series-To-Text), a domain-agnostic neural framework for time-series narration, that couples the representation of essential time-series elements in the form of a dense knowledge graph and the translation of said knowledge graph into rich and fluent narratives through the transfer learning capabilities of PLMs (Pre-trained Language Models). T\textsuperscript{3}'s design primarily addresses the challenge that lies in building a neural framework in the complete paucity of annotated training data for time-series. The design incorporates knowledge graphs as an intermediary for the representation of essential time-series elements which can be linearized for textual translation. To the best of our knowledge, \name is the first investigation of the use of neural strategies for time-series narration. Through extensive evaluations, we show that \name can improve the lexical diversity of the generated narratives by up to 65.38\% while still maintaining grammatical integrity. The practicality and deployability of \name is further validated through an expert review ($n=21$) where 76.2\% of participating experts wary of auto-generated narratives favored \name as a deployable system for time-series narration due to its richer narratives. Our code-base, models, and datasets, with detailed instructions for reproducibility is $publicly$ hosted\footnote{\url{https://github.com/Mandar-Sharma/TCube}}.
\end{abstract}

\begin{IEEEkeywords}
time-series analysis, time-series-to-text, data-to-text, pre-trained language models, natural language generation
\end{IEEEkeywords}

\section{Introduction}
Real-world data is often temporal in nature. From the global outbreaks of infectious diseases to the prices of stocks, all chronologically recorded data takes the form of a time-series. Thus, its mining and analysis has been of significant interest to the scientific community \cite{ts_mining}. Time-series narration aims to portray the discerning characteristics of a time-series obtained from such analysis through a textual narrative. The efficacy of narratives as an aid to data comprehension has been validated through studies in digital libraries \cite{latif} as well as causal networks \cite{causal}. Petre, in his advocacy for the importance of textual representations of data \cite{petre}, humorously notes, \say{A picture is worth a thousand words - isn't it? And hence graphical representation is by its nature universally superior to text - isn't it? Why then isn't the anecdote itself expressed graphically?}. 

\begin{figure*}[t]
    \centering
    \includegraphics[width=\linewidth]{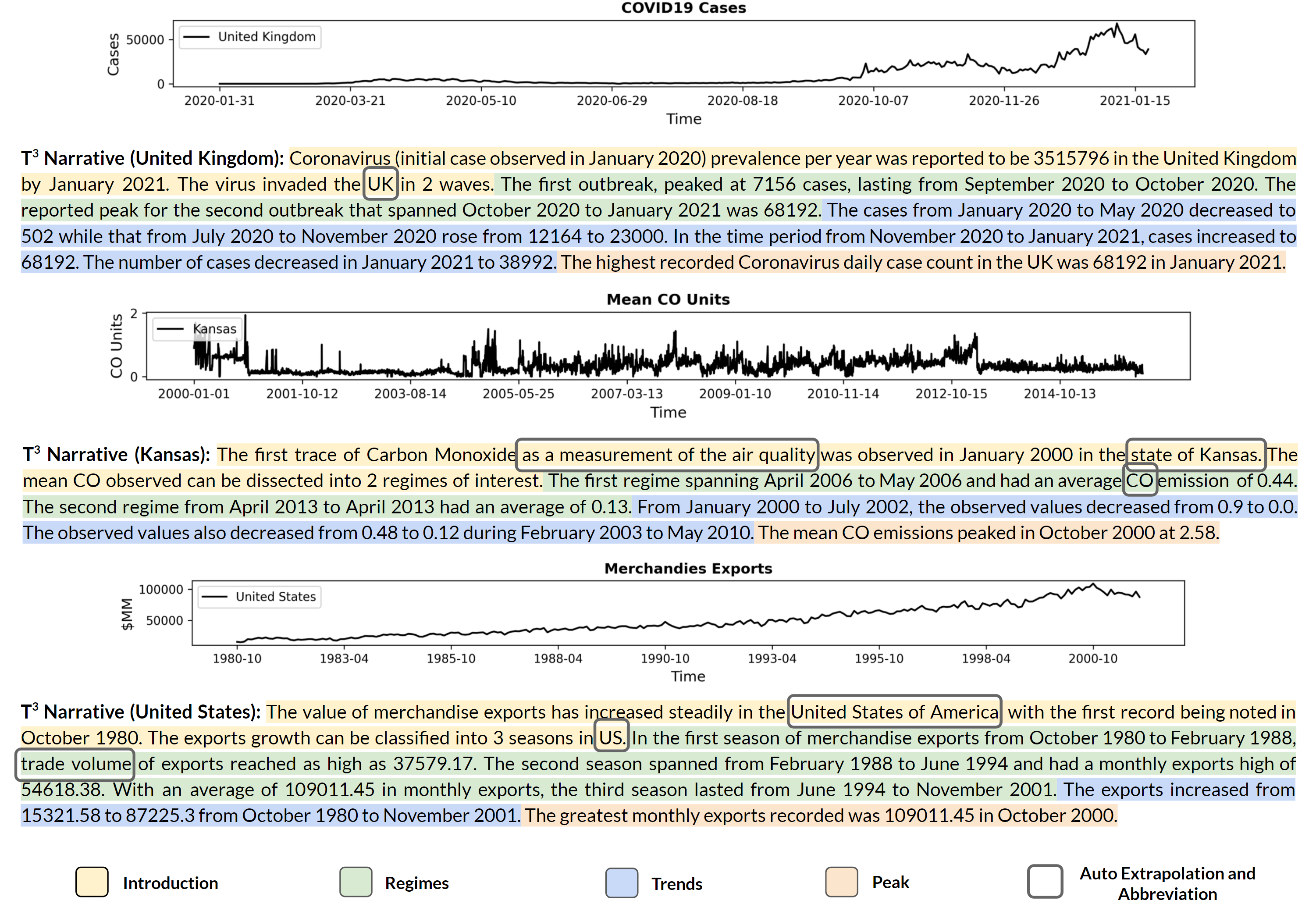}
    \caption{Sample narratives generated by \name for the United Kingdom COVID19, Kansas CO pollution, and United States merchandise exports datasets.}
    \label{fig:teaser}
\end{figure*}

Time-series narration falls under the umbrella of data-to-text, a sub-field of NLG (Natural Language Generation) that aims to produce meaningful and coherent textual descriptions of non-linguistic data \cite{reiter_dale, gatt_krahmer}. Although data-to-text has garnered significant interest over the years, recent efforts for textual description of data have been focused on either tabular data \cite{table_to_text, content_planning, hierarchical_d2t} or graph data \cite{triple_to_text, graph2text}. The attention that these data types have garnered simultaneously highlight the two key challenges for time-series data. The first being in the design and training of such a system in the paucity of \say{gold} datasets and the second in its evaluation standards.
\begin{itemize}
    \item End-to-end models for data-to-text generation showcase learning a direct input-output mapping from data to text \cite{wikibio, end-to-end} through the use of annotated datasets such as WikiBio \cite{wikibio} and E2E \cite{e2e} for tabular data and WebNLG and DART \cite{webnlg, dart} for RDF (Resource Description Framework) triples \cite{rdf}. In both tabular data and RDF triples, the information to be presented in the narrative is present in the data itself and is copied to the output token - making end-to-end learning possible. In contrast, time-series requires further processing for the discovery of underlying patterns to be narrated. Thus, due to the inherent numerical and continuous nature of time-series, one needs to consider time-series as a whole rather than a sum of its individual constituents. Thus, one would have to either follow the traditional modular pipeline architecture \cite{reiter_dale} where non-linguistic data is transformed into text through several intermediate steps, or formulate a novel approach suited to time-series data altogether.
   \item  The \say{gold} narratives in the aforementioned datasets offers a common ground for automated evaluation of competing frameworks on the basis of word-based metrics such as BLEU \cite{bleu} and its variants \cite{rouge, meteor, chrf}. Thus, there are domain-familiar metrics present to showcase how one framework can perform better than another. For time-series data, without human annotations corresponding to the data, automated evaluation through said word-based metrics is not possible.
\end{itemize}
 
As will be discussed in the related works section, there have been several previous efforts for time-series narration. Although these pioneering efforts have laid significant groundwork for this field, the recent work in time-series narration falls short in two crucial areas: First, they are domain-specific, modeled specifically for use in fields such as meteorology, intensive care, health monitoring and so on. Second, the proposed systems have not actualized the recent advances in language processing, rather, relying on the traditional pipeline architecture. Graefe et. al. \cite{readers} note \say{news consumers get more pleasure out of reading human-written as opposed to computer-written content}. Thus, these template-based narratives can be met with a dismissive response by its users due to its seemingly mechanical nature - we further elaborate on this in our expert review section.

To address these challenges, we present \name: Timeseries-To-Text, which stands out from previous forays in this task through a) its domain-agnostic nature and b) its coupling of dense knowledge graph based representation of essential time-series elements and the translation of said knowledge graph into rich and fluent narratives through the transfer-learning capabilities of large PLMs (Pre-trained Language Models) fine-tuned to this specific task - tackling the paucity of annotated data. Figure \ref{fig:teaser} highlights the diversity in the narratives generated by T\textsuperscript{3} along with the automatic extrapolations and abbreviations deduced by the language models. The terms `United Kingdom', `United States', and `Carbon Monoxide' are automatically abbreviated to `UK', `US', and `CO' respectively. Similarly, the system extrapolates information such as adding `as a measurement of air quality' when mentioning carbon monoxide values, adding `the state of' to Kansas, and introducing the term `trade volume' when describing export values. Our contributions are summarized as follows:
\begin{itemize}
    \item To the best of our knowledge, \name is the first foray into neural time-series narration. Our rigorous evaluations across multi-domain datasets showcases that \name consistently produces 65.38\% more diverse narratives with the same grammatical integrity as the existing baselines.
    \item Through an expert review $(n=21)$, we validate the performance, practicality, and linguistic superiority of $T^{3}$. 76.2\% of participating experts who were wary of auto-generated narratives favored \name as a deployable system as compared to existing baselines.
    \item We benchmark the performance of several time-series segmentation and regime-shift detection algorithms as well as prominent PLMs for outlining the best approach to a domain-agnostic time-series narration framework.
    \item Our code-base, pre-trained models, the datasets used, along with a detailed notebook guide for reproducibility are made public\textsuperscript{1}.
\end{itemize}

\section{Narratives: Good, Bad, and Boring}
Textual narratives are swiftly becoming important components of visualization systems, either as a way to generate data insights to accompany visualizations \cite{endert-vast} or to structure visualizations for better communication \cite{fromendert}. Research into what makes an effective narrative is still in its infancy and is necessarily tied to the underlying analytical task and domain. For temporal data, we identify the following crucial facets:

\noindent \textbf{Level of detail:} Should the narrative capture an executive summary or provide in-depth access to the underlying data?

\noindent \textbf{Language diversity:} Greater diversity in language prevents monotony but could detract from conveying key messages and conclusions. Lower diversity, on the other hand, supports comparison of different narratives, but leads to \say{glossing over} by analysts - defeating the very purpose of these narratives.

\noindent \textbf{Verbalizing numbers:} The verbalization of quantitative or probabilistic data (using Kent's words of estimative probability \cite{kent} or the NIC/ Mercyhurst standardization) and trends is considered important in specific domains (such as intelligence analysis \cite{heuer}), however, other applications argue for direct access to the original numeric information.

\noindent \textbf{Human performance aspects:} Understanding the characteristics of narratives that lead to improved human performance is an ongoing research problem \cite{metoyer2018coupling}. Narratives provide increased comprehension, interest, and engagement and are known to  contribute \say{distinct cognitive pathways of comprehension} with increased recall, ease of comprehension, and shorter reading times \cite{dahlstrom2014}. Conversely, the challenge of the written word implies slowness and error-prone behavior due to short-term memory limits. 

In essence, successful narrative research requires a standardization of both the generation and evaluation space, and an understanding of how a narrative fits into the larger comprehension process of the analyst. As an example, a \say{bad} narrative for a fictional monthly sales-volume dataset, in the form of \say{\textit{The sales numbers for January 2019 were 1500 while the sales numbers for February 2019 were 2000. Similarly, the sales numbers for ...}}, falls to meet all the above criterion: it is lexically repetitive, portrays no information about the data that would have been difficult to discern visually, and presents the numbers as-is with no verbalization.

\section{Related Work}
While some of the earliest work on time-series narration can be traced back to 1994 with the Forecast Generator (FOG) \cite{fog}, a framework for generating bilingual (English/French) textual summaries of weather forecasts, in the recent decades, Ehud Reiter's research group has laid significant groundwork for this domain. Their SUMTIME-MOUSAM project \cite{reiter_sumtime} generates short textual summaries of weather forecasts and SUMTIME-TURBINE \cite{sumtime_turbine} generates the same for sensor readings from a gas turbine. The design of these SUMTIME systems highlights the importance of domain expertise in relaying the information embedded in a raw time-series in a manner relevant to the end user. Following this, their SUMTIME project was extended to SUMTIME-NEONATE \cite{SUMTIME_neonatal}, which generates textual summaries of time-series data intended to aid medical professionals in monitoring infants in neonatal intensive care units. In 2003 \cite{reiter_gricean}, the authors highlight the use of Gricean maxims of cooperative communication \cite{grice} for the selection of the most crucial information to be relayed to the end user. The authors further investigate the impact of word choice in textual summarization by avoiding words specific to one idiolect and words whose meanings varied in different idiolects \cite{reiter_idio}.

Kacprzyk et. al. \cite{kacprzyk} propose the use of Zadeh’s calculus of linguistically quantified propositions with varying t-norms to summarize time-series segmented with Piece-wise Linear Approximations. Castillo-Ortega et. al. \cite{castillo}, propose linguistic summarization of time-series based on the hierarchical structure of time. The multiple candidate summaries are evaluated with a multi-objective evolutionary algorithm. In the physiological domain, Banaee et. al. \cite{banaee} propose a system to summarize the data streams from health monitoring systems in a clinician and patient centric manner. Dubey et. al. \cite{dubey} propose the use of Case-based Reasoning from records of previous summaries to summarize weather reports. 

Thus, there has been significant investigation into this domain. However, the research emphasis has heavily been in the identification of the information to relay to the end user rather than relaying the information in a manner engaging to the end user - having the narratives themselves be rich and fluent. The textual output of the above mentioned systems follow the traditional modular pipeline architecture of Reiter and Dale \cite{reiter_dale}. Commercial services such as The Automatic Statistician\footnote{\url{https://automaticstatistician.com/}} and Narrative Science\footnote{\url{https://narrativescience.com/}} offer data summarization through visualization and narratives. Although their technology and code is proprietary, a perusal through offered samples\footnote{\url{https://automaticstatistician.com/examples/}} for time-series summarization hints towards templated generation where variables from analysis are plugged into preset templates.

\section{Preliminaries}
In this section we outline some necessary background in time-series segmentation, detecting shifting regimes, and PLMs, as a foundation for T\textsuperscript{3}'s architecture.
\subsection{Segmentation}
Given a time-series $T$ of length $n$, a segmentation of $T$ contains a set of distinct temporal cut-points $S=\{c_1,c_2,..,c_{k}\}$ corresponding to $k$ straight lines where $k << n$ \cite{muralidhar}. The segmentation approach can be limited by the number of segments $k$ produced, or by a predefined threshold for segment-wise or cumulative error. As time-series of varying types and lengths need to be approximated with varying number of segments, we evaluate the following candidate segmentation algorithms based on a preset error threshold to promote domain-agnosticism.

\noindent \textbf{Sliding Windows}: The data points from a time-series are added to a sliding window until the maximum approximation error is met and a segment is formed. This process repeats with the window starting from the next data point.

\noindent \textbf{Bottom-Up}: The algorithm starts with the finest approximation such that a time-series of length $n$ is approximated by $\frac{n}{2}$ segments. The algorithm iteratively merges the lowest cost adjacent segments until the stopping criteria is met. 

\noindent \textbf{SWAB}: An acronym for the integration of Sliding Windows and Bottom-Up, SWAB \cite{swab} first defines an initial buffer $w$ on which Bottom-Up is performed. The first segment from $w$ is reported and the corresponding data points are removed from it. Remaining points from the series are read into $w$ till the linear fit on it reaches an error threshold. This process is repeated until the buffer $w$ reaches the end of the time-series.

\subsection{Regime-shifts} 
Regime shift or switching refers to changes in the state or structure of a time-series. For domain-agnosticism, we require the shift-detection algorithms to be unsupervised, universal approximators, and input length invariant. Thus, based on these criterion, we evaluate the following candidates:

\noindent \textbf{Rrepresentation Learning}: Franceschi et. al.'s \cite{franceschi} unsupervised representation learning algorithm, hereby noted as \say{\textit{RL}}, learns representations of time-series elements using an encoder architecture based on causal dilated convolutions with a triplet loss arrangement that employs time-based negative sampling.

\noindent \textbf{Matrix Profile}: The Matrix Profile \cite{matrixprofile, matrixprofile_code} is a multi-purpose annotation (profile) of a time-series $T$ where the $i^{th}$ location on the profile records the distance of the sub-sequence in $T$ at the $i^{th}$ location to its nearest neighbor.

\begin{figure*}[t]
    \centering
    \includegraphics[width=\linewidth]{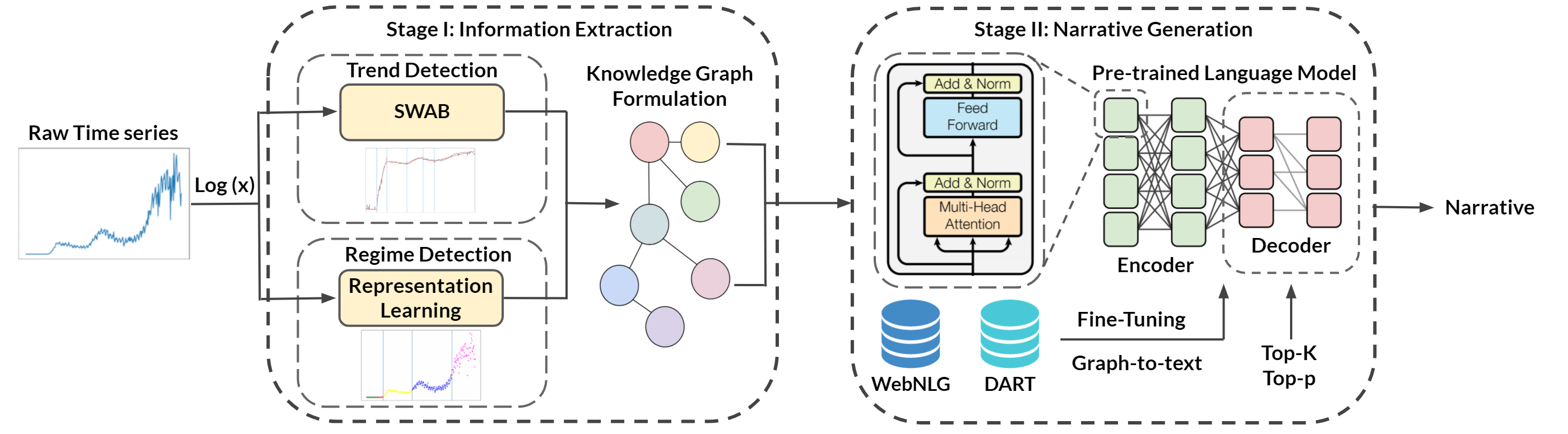}
    \caption{The two stage \name framework: In Stage I, the system extracts trends, regimes, and peaks from the input time-series which is formulated into a knowledge graph. In Stage II, a PLM fine-tuned for graph-to-text generation generates the narrative from the input graph.}
    \label{fig:arch}
\end{figure*}

\subsection{Pre-trained Language Models}
Transfer learning in language processing has been democratized and made universal with the advent of PLMs \cite{plm_survey} which share the multi-headed attention core architecture of transformers \cite{transformers}. Transfer learning, in the context of PLMs, is essentially the adaptation of these massive language models to downstream tasks such as data-to-text, question answering, summarization and much more via a fine-tuning process on task-specific data. Through the effors of Thomas Wolf et. al. \cite{huggingface}, second-generation seq2seq PLMs such as Google's T5 \cite{t5} and Facebook's BART \cite{bart} and auto-regressive PLMs such as Open-AI's GPT-2 \cite{gpt2} and many more have been made accessible to the larger community.

The motivation behind using PLMs for this task not only stems from the fact that they lead the benchmark for a multitude of downstream language processing tasks \cite{t5_beats_all} but also due to the evidence that PLMs, due to their apparent acquisition of worldly knowledge \cite{plms_opengraph}, in some cases refuse to generate false outputs even when the input to the system is corrupted \cite{graph2text}. As Open AI's GPT-3 \cite{gpt3} has not been released for public access at the time of publication of this paper, we have not been able to incorporate it into our experiments.

\subsection{Decoding Strategies}
The PLMs we intend to investigate---viz. Open-AI's GPT-2, Facebook's BART, and Google's T5---though differing in their architectures and training strategies, share an auto-regressive decoder. Auto-regressive language generation is based on the assumption that the probability distribution of a sequence of words can be decomposed into the product of conditional next word distributions. If $W_0$ be the initial context word sequence and $T$ be the length of the sequence to be generated, then the probability distribution can be defined as:
\begin{equation}
    P(w_{1:T} | W_{0}) = \prod_{t=1}^{T} P(w_{t} | w_{1:t-1}, W_{0})
\end{equation}
\noindent \textbf{Basic Sampling}: This strategy is based on randomly picking a word $w_t$ based on its conditional probability distribution $w_t \sim P(w|w_{1:t-1})$. Thus, the next word in the sequence is chosen based on its conditional probability of occurrence.

\noindent \textbf{Top-K Sampling}: In top-K sampling \cite{topk}, the $K$ words most likely to occur next in the sequence are chosen and the probability mass is redistributed among these $K$ words. This leads to a more \say{human-like} text generation.

\noindent \textbf{Top-p Sampling}: Top-p sampling \cite{topp}, also known as nucleus sampling, addresses a core issue in top-K sampling. Since top-K re-distributes the probability mass among the top $K$ chosen words, it has the potential to break down in particularly sharp or flat distributions. If a distribution is sharp, the limit on the selection of just $K$ words can lead to insensible text generation. On the other hand, for flat distributions, the limit prevents the generation from being diverse. Thus, instead of limiting the sampling space to $K$ words, top-p samples from the smallest possible set of words whose cumulative probability exceeds a predefined probability $p$.

\section{\name Framework}
\subsection{The Architecture}

The two-stage design of T\textsuperscript{3}, as illustrated in figure \ref{fig:arch}, is motivated by the need to produce rich and fluent narratives of time-series data with the least-possible human intervention. Subsections VI-A and VI-B highlight thorough experimentation that motivate the specific choices for the segmentation and regime-shift detection algorithms for \name while subsections VI-C and VI-D highlight the same for our choice of PLMs.

\begin{figure}[!h]
    \centering
    \includegraphics[scale=0.5]{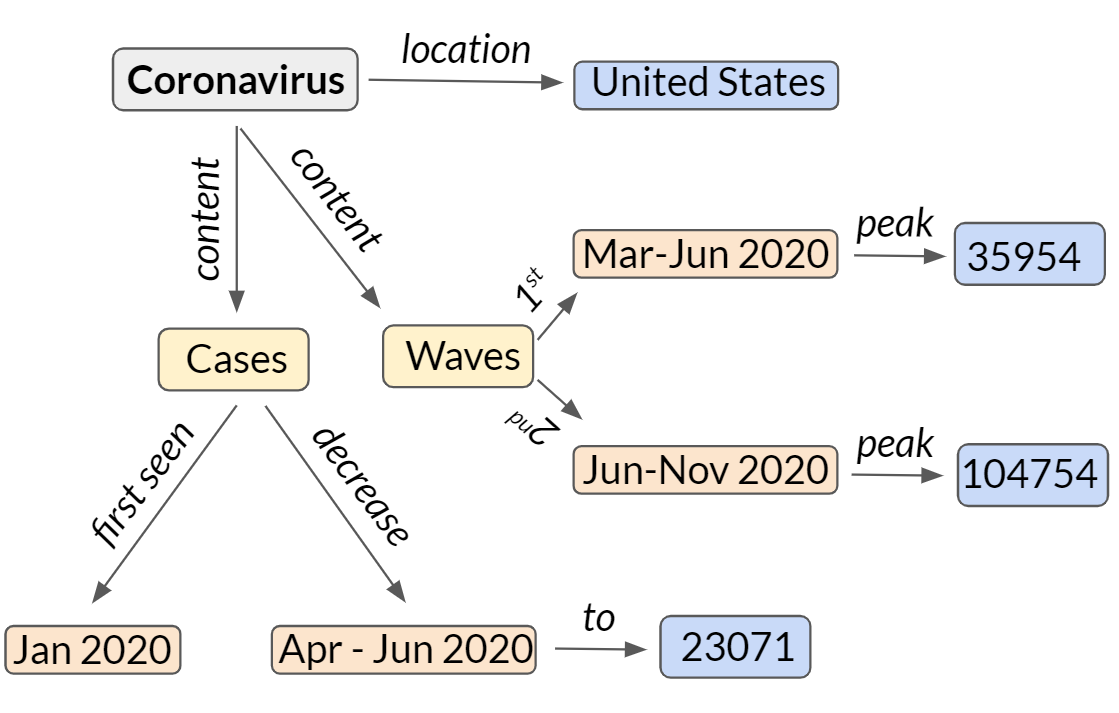}
    \caption{Sample \name knowledge graph (curtailed) for U.S. COVID19 Cases.}
    \label{fig:samplekg}
\end{figure}

\noindent \newline \textbf{Stage I}: The time-series is first log-transformed to approximately conform the data to normality before information extraction. This log-transformed series is segmented into $k$ linear segments where the individual slopes of these $k$ segments indicates the trends followed by the data in their respective intervals. Simultaneously, sequential data-points with similar properties are clustered together based on their learned representations. These clusters represent changing regimes in the dataset. The above time-series characteristics are encoded into a RDF-based knowledge graph. Figure \ref{fig:samplekg} illustrates a sample knowledge graph (curtailed) as extracted from T\textsuperscript{3}'s first stage for the United States COVID19 time-series.

\noindent \textbf{Stage II}: Anterior to T\textsuperscript{3}'s execution, the PLMs are fine-tuned with both WebNLG and DART datasets for graph-to-text translation. The knowledge graph from Stage I is thus translated into a rich and descriptive narrative by these PLMs using sampling techniques for strategic language generation.

\subsection{Datasets}
\noindent \textbf{Time-series:} To promote domain-agnosticism, the datasets used for evaluating \name are drawn from five different fields - COVID19\footnote{\url{https://ourworldindata.org/}}, Direction of Trade Statistics\footnote{\url{https://data.imf.org/}}, Carbon Monoxide Pollution\footnote{\url{https://data.world/data-society/}}, World Population\textsuperscript{4}, and Climate Change\textsuperscript{4}. Based on the amount and consistency of the data, we consider the same ten countries (United States, India, Brazil, Russia, United Kingdom, France, Spain, Italy, Turkey, and Germany) across these datasets. The CO (Carbon Monoxide) units, however, are extracted for the U.S. states with EPA state codes 1 through 10. Table \ref{table:1} provides a brief statistical summary of these datasets.

\noindent \newline \textbf{Fine-tuning:} RDF-based datasets WebNLG v3.0 and DART v1.1 are used for fine-tuning the PLMs in T\textsuperscript{3}. Table \ref{table:1} briefly summarizes the statistics of these datasets where $N_{x}$ represents the number of samples for $x \in \{train, dev, test\}$ and $V$, $WSR$, and $SSR$ represent the vocabulary size, words per SR (Surface Realization), and sentence per SR respectively.

\subsection{Fine-tuning, Training, and Decoding Specifications}
Tokens $<$X$>$ where $X \in \{H,R,T\}$ are appended to the start of the Head (subject), Relationship (predicate), and Tail (object) entities of each RDF triple. The Adam optimizer \cite{adam} with a linearly decreasing learning rate is used to fine-tune the PLMs with learning rates initially set to 3e-5 for T5 and BART and 5e-4 for GPT-2. For uniformity, the maximum token lengths for all PLMs are set to their default maximum (512) with a batch size of 4. For strategic decoding, based on the average length ($\sim$100 words) and the average number of unique words ($\sim$50) present in the generated narratives we set $k$ as 50. Similarly, based on popular practice, we set $p$ as 92\%.

\vspace{-0.25cm}
\begin{table}[h]
\centering
\caption{Datasets statistics.}
\label{table:1}
\begin{tabular}{@{}cccc@{}}
\toprule
\multicolumn{1}{l}{} & \multicolumn{1}{l}{$N$} & \multicolumn{1}{l}{$\mu$} & \multicolumn{1}{l}{$\sigma$} \\ \midrule
COVID19 Cases            & 351                     & 1.75e4                    & 1.95e4                       \\
DOTS Exports (MM)                 & 254                     & 1.35e4                    & 6.27e3                       \\
U.S. CO Pollution (Units)         & 4722                    & 0.39                      & 0.22                         \\
World Population     & 22                      & 8.02e7                    & 4.82e7                       \\
Global Temperature (\textdegree{C})  & 3166                    & 8.15                      & 6.91                         \\ 
\end{tabular}
\begin{tabular}{@{}lcccccc@{}}
\toprule
      & $N_{train}$ & $N_{dev}$ & $N_{test}$ & $V$     & $WSR$  & $SSR$ \\ \midrule
WebNLG & 35426   & 4464   & 5150   & 8000  & 22.5 & 1.4 \\
DART   & 62659   & 6980   & 12551  & 33200 & 21.6 & 1.5 \\ \bottomrule
\end{tabular}
\end{table}
\vspace{-0.2cm}

\section{Experiments}
Through our experimentation, we seek to address three core questions regarding the design and need for a domain-agnostic time-series narration framework:
\begin{itemize}
    \item For the design of a domain-agnostic narration framework, how do we choose among the prominent time-series analysis tools at our disposal? (Sections A and B)
    \item How do state-of-the-art language models stack up against each other for the task of translating knowledge graphs to natural language? (Section C)
    \item Does \name deliver richer and more diverse narratives as compared to traditional approaches? Does \name hallucinate? Would domain experts find this favorable and/or practical? (Sections D and VII)
\end{itemize}

\begin{table*}[t]
\centering
\caption{Comparison of time-series segmentation algorithms based on total $SSE$ of residuals and $r^2$ fit across datasets.}
\label{table:3}
\begin{tabular}{lcccccccccc}
\hline
               & \multicolumn{2}{c}{COVID19 Cases}          & \multicolumn{2}{c}{DOTS Exports}           & \multicolumn{2}{c}{U.S. CO Pollution}      & \multicolumn{2}{c}{World Population}       & \multicolumn{2}{c}{Global Temperature} \\ \hline
               & \multicolumn{1}{c|}{$SSE$} & $r^2$         & \multicolumn{1}{c|}{$SSE$} & $r^2$         & \multicolumn{1}{c|}{$SSE$} & $r^2$         & \multicolumn{1}{c|}{$SSE$} & $r^2$         & \multicolumn{1}{c|}{$SSE$}   & $r^2$   \\ \hline
Sliding Window & 300.61                     & 0.12          & 6.91                       & 0.08          & 73.91                      & 0.15          & 0.92                       & 0.61          & 838.35                       & 0.16    \\
Bottom-Up      & \textbf{27.07}             & 0.13          & 4.98                       & 0.07          & 67.53                      & 0.16          & 0.75                       & \textbf{0.64} & 67.36                        & 0.16    \\
SWAB           & 27.16                      & \textbf{0.14} & \textbf{4.90}              & \textbf{0.08} & \textbf{67.11}             & \textbf{0.16} & \textbf{0.11}              & 0.37          & \textbf{65.21}               & \textbf{0.17}    \\ \hline
\end{tabular}
\end{table*}

\subsection{Trend Detection}
In order to evaluate our candidate segmentation algorithms, we must first determine the right value of allowable maximum linear-fit error appropriate for our datasets. The evaluation of the total SSE (Sum Squared of Errors) of residuals vs $k$ (the numbers of segments produced), as a function of the maxmimum linear-fit error, hints at 2.75 as a potential error \say{sweet spot}. The figure below presents this analysis for the U.S. COVID19 dataset - the left marker indicates the trade-off point between the total SSE and $k$ while the right marker indicates the point where both total SSE and $k$ stabilize.

\vspace{-0.2cm}
\begin{figure}[h]
    \centering
    \includegraphics[width=\linewidth]{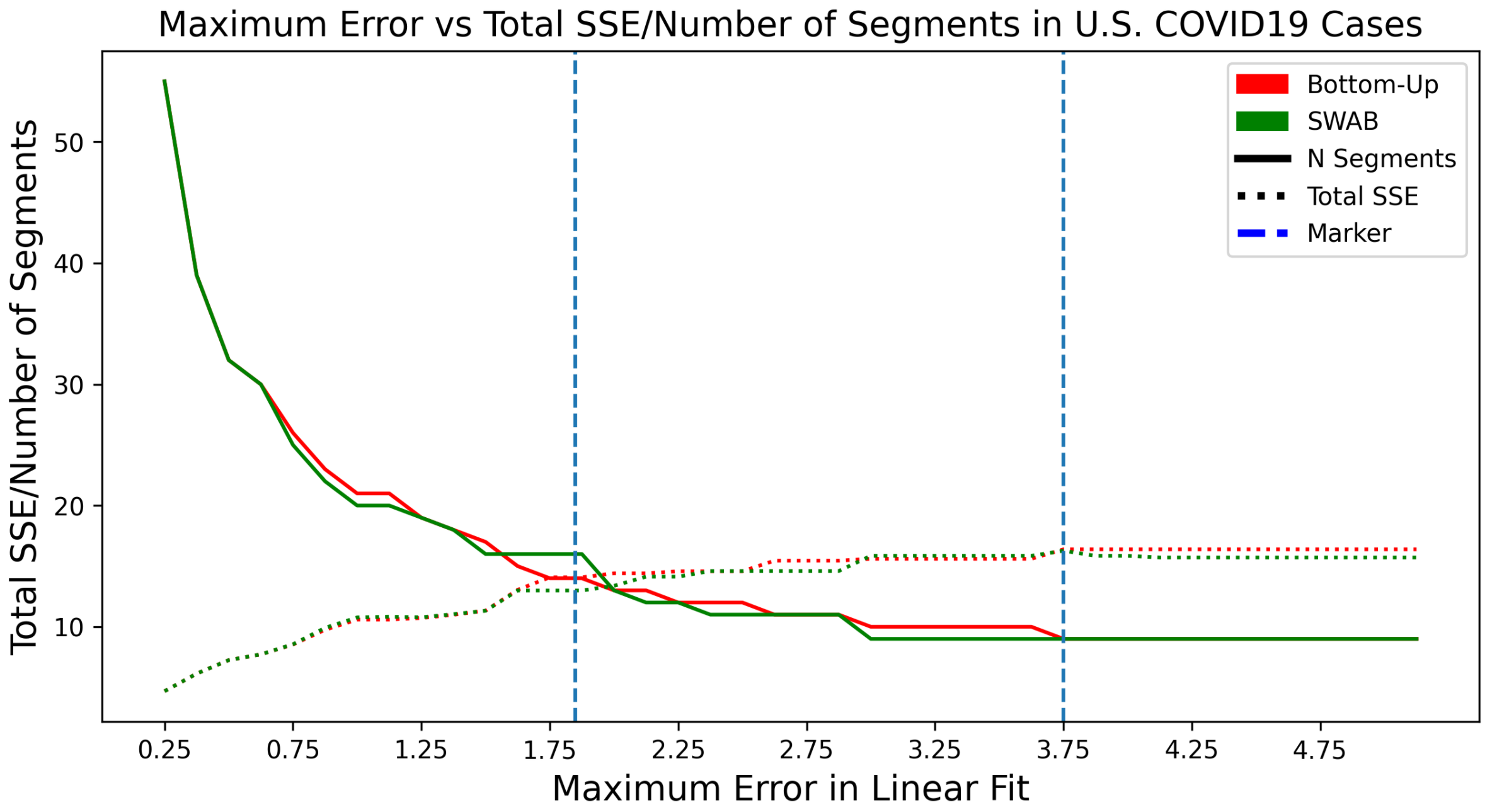}
    %\caption{Total SSE and the number of segments as a function of max-error.}
    %\label{fig:covid}
\end{figure}

\vspace{-0.25cm}
Table \ref{table:3} outlines the performance of the selected segmentation algorithms across our datasets with the maximum linear-fit threshold set to 2.75. We observe that SWAB consistently performs the best in terms of both the $r^2$ goodness-of-fit and $SSE$, making it the segmentation algorithm of choice for T\textsuperscript{3}.

\vspace{-0.25cm}
\begin{figure}[h]
    \centering
    \includegraphics[width=\linewidth]{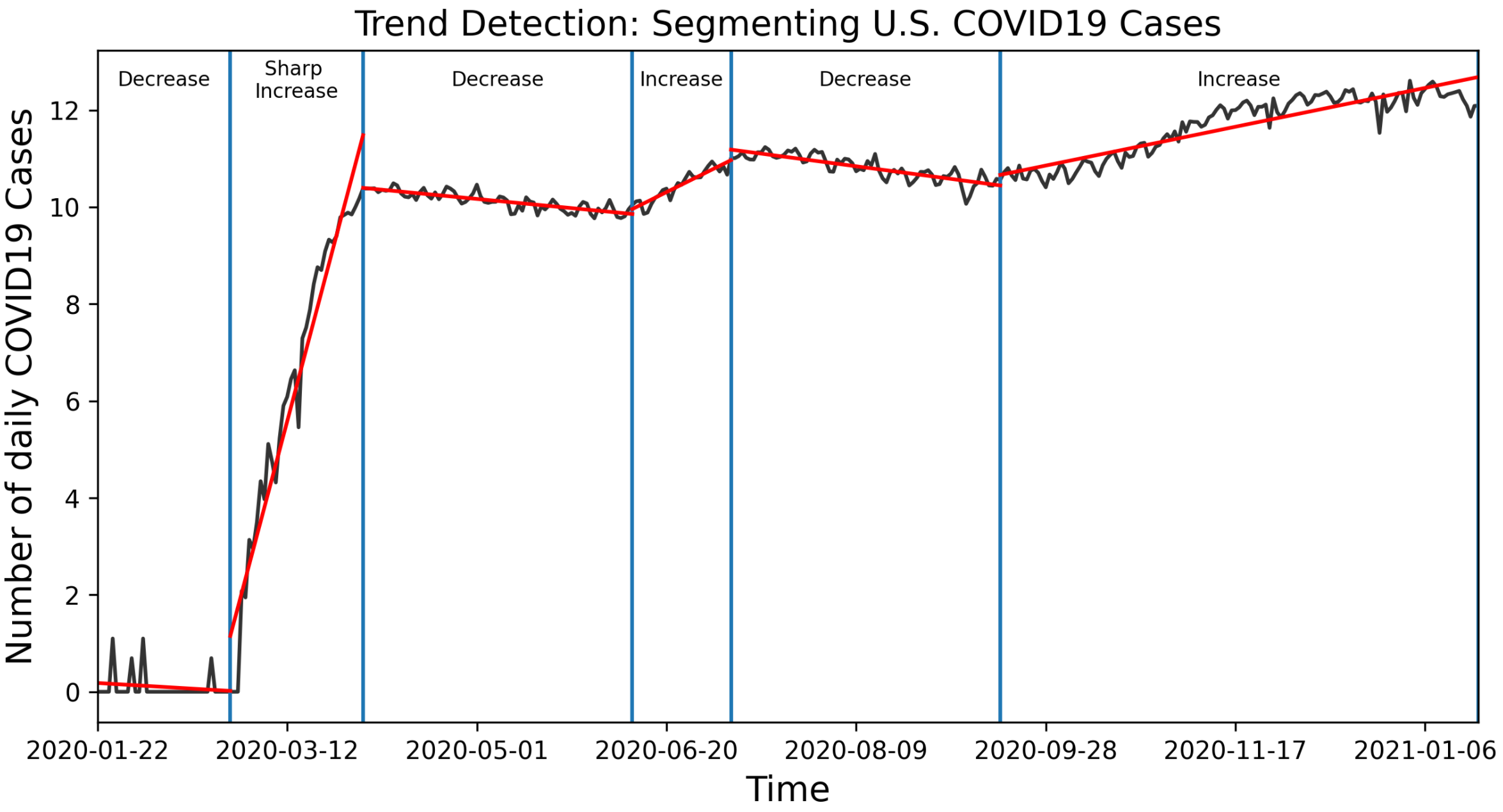}
    %\caption{Trend detection in United States COVID19 Cases}
    %\label{fig:trend}
\end{figure}

\vspace{-0.2cm}
Out of the $k$ segments produced for each time-series, if the slope of $k_{i-1}^{th}$ segment follows that of the $k_{i}^{th}$ segment, we rearrange them as a single segment for continuity. This is illustrated in the figure above for the U.S. COVID19 time-series where the original $k$ segments are consolidated based on their slopes to 6 long segments ($k>6$) that indicate the core trends followed by the time-series over significant time-spans.

\subsection{Regime Shift Detection}
For the evaluation of our candidate regime-shift detection algorithms, we force these algorithms to produce a known number of regime shifts validated through visual interpretation of the data - regime shifts in COVID19 cases should correspond to waves of outbreak, as illustrated in the figure below, whereas those in DOTS Exports should correlate to inflation or deflation in the economy.

\vspace{-0.2cm}
\begin{figure}[h]
    \centering
    \includegraphics[width=\linewidth]{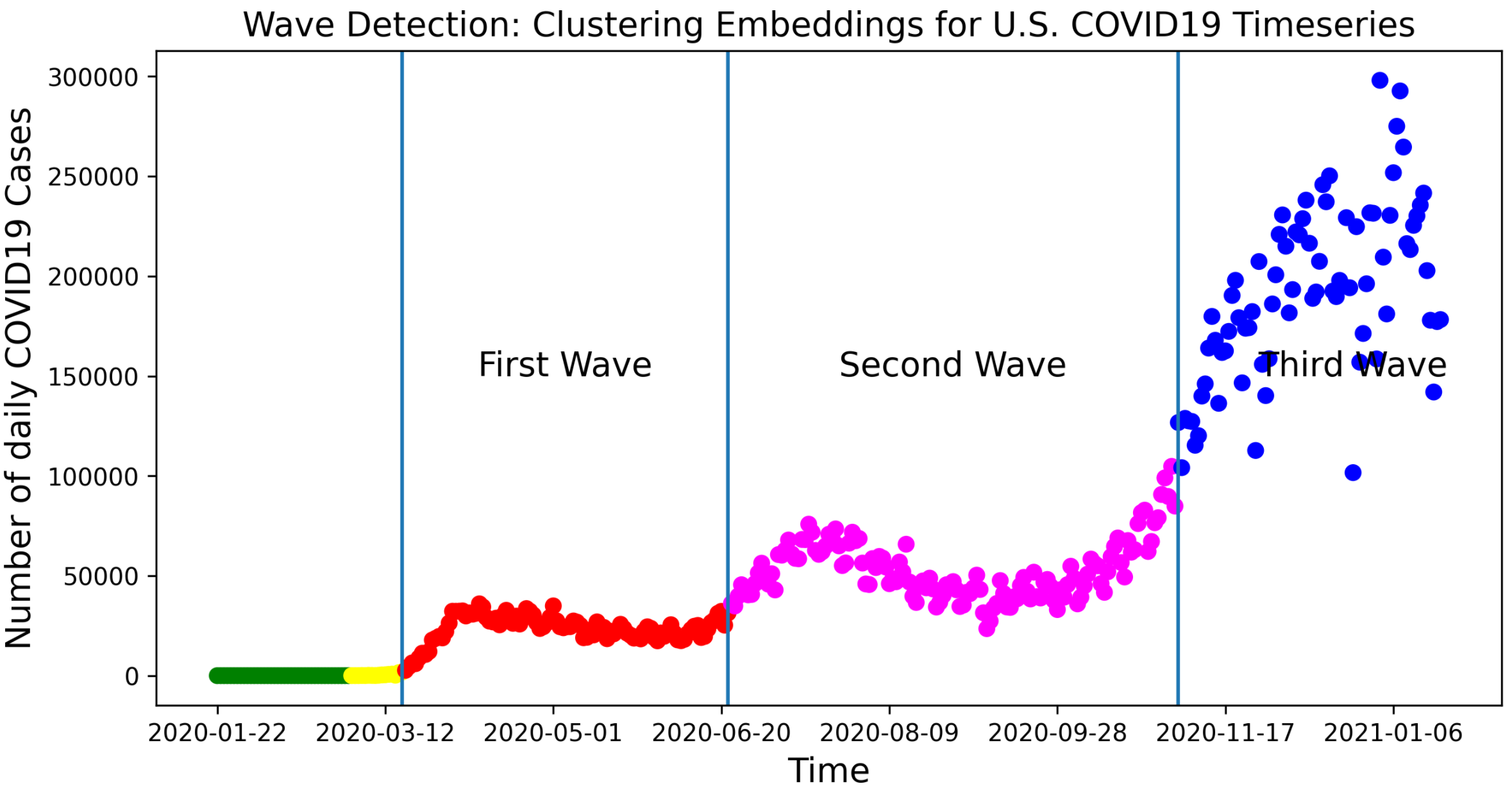}
    % \caption{Wave Detection: Regime-shift detection in U.S. COVID19 time-series with Representation Learning}
    % \label{fig:waves}
\end{figure}

\vspace{-0.25cm}
Table \ref{table:4} outlines the performance of Matrix Profile and RL across our datasets based on the standard deviations ($\sigma$) of the formed regimes. Our evaluations lead us to conclude that the performance of Matrix Profile and RL are on-par and vary based on the individual dataset. In our implementation, an RL instance trained on the COVID19 dataset showcases high cross-domain transferability when applied to other series in our catalog. The Matrix Profile, however, requires a window-size definition prior to its execution which varies based on the input time-series. The tendency of RL to favor automation makes it the regime-shift detection algorithm of choice for T\textsuperscript{3}.

\vspace{-0.3cm}
\begin{table}[h]
\centering
\caption{Comparison of regime shift detection algorithms based on $\sigma$.}
\label{table:4}
\begin{tabular}{lcc}
\hline
                   & Matrix Profile ($\sigma$) & RL ($\sigma$)             \\ \hline
COVID19 Cases      & \textbf{7.29}  & 9.43           \\
DOTS Exports       & \textbf{8.68}  & 8.98           \\
U.S. CO Pollution  & \textbf{2.35}  & 2.41           \\
World Population   & -              & \textbf{15.26} \\
Global Temperature & 2.20           & \textbf{2.19}  \\ \hline
\end{tabular}
\end{table}

\subsection{Graph-to-text Translation}
\begin{table*}[t]
\centering
\caption{Evaluation of PLMs on graph-to-text translation on the WebNLG dataset, DART dataset, and their combination.}
\label{table:5}
\begin{tabular}{@{}lcccccccccccc@{}}
\toprule
Dataset    & \multicolumn{4}{c}{WebNLG}                                                     & \multicolumn{4}{c}{DART}                                                       & \multicolumn{4}{c}{WebNLG + DART}                                 \\ \midrule
Model      & BLEU           & ROUGE          & METEOR         & \multicolumn{1}{c|}{chrF++} & BLEU           & ROUGE          & METEOR         & \multicolumn{1}{c|}{chrF++} & BLEU           & ROUGE          & METEOR         & chrF++         \\ \midrule
GPT-2      & 14.2           & 4.28           & 20.22          & 37.13                       & 15.56          & 5.23           & 21.79          & 37.68                       & 18.65          & 7.54           & 23.61          & 39.22          \\
BART Base  & 32.13          & 51.81          & 33.49          & 59.08                       & 33.77          & 54.27          & 35.86          & 61.15                       & 37.89          & 58.22          & 37.80          & 64.52          \\
BART Large & 32.04          & 51.10          & 34.68          & 59.98                       & 34.75          & 55.32          & 36.47          & 61.75                       & 38.36          & 58.18          & 38.15          & 64.82          \\
T5 Small   & 33.94          & 56.46          & 35.4           & 61.56                       & 34.52          & 55.96          & 36.33          & 61.74                       & 38.52          & 59.05          & 38.21          & 65.06          \\
T5 Base    & \textbf{36.75} & \textbf{57.76} & \textbf{37.25} & \textbf{64.17}              & \textbf{36.40} & \textbf{57.00} & \textbf{37.44} & \textbf{63.23}              & \textbf{39.88} & \textbf{59.71} & \textbf{38.91} & \textbf{65.95} \\ \bottomrule
\end{tabular}
\end{table*}

\begin{table*}[t]
\centering
\caption{Comparison of the performance of \name with that of Templated Generation based on language evaluation metrics.}
\label{table:6}
\begin{tabular}{@{}lccccccccccccccc@{}}
\toprule
\multicolumn{1}{c}{} & \multicolumn{3}{c}{COVID19 Cases}                       & \multicolumn{3}{c}{DOTS Exports}                        & \multicolumn{3}{c}{U.S. CO Pollution}                   & \multicolumn{3}{c}{World Population}                    & \multicolumn{3}{c}{Global Temperature}      \\ \midrule
\multicolumn{1}{c}{} & RE             & TTR           & \multicolumn{1}{c|}{G} & RE             & TTR           & \multicolumn{1}{c|}{G} & RE             & TTR           & \multicolumn{1}{c|}{G} & RE             & TTR           & \multicolumn{1}{c|}{G} & RE             & TTR           & G          \\ \midrule
Templated Generation & 17.79          & 0.26          & \textbf{1}             & 54.73          & 0.45          & \textbf{1}             & 64.34          & 0.22          & \textbf{1}             & 66.28          & 0.46          & \textbf{1}             & 55.24          & 0.37          & \textbf{1} \\
\name with $T5$           & 64.48          & 0.31          & 0.99                   & 67.54          & 0.47          & \textbf{1}             & 69.22          & 0.28          & 0.99                   & 69.56          & 0.49          & \textbf{1}             & \textbf{67.45} & 0.39          & 0.99       \\
\hspace{1cm} $T5_{top-K}$           & 65.67          & 0.38          & 0.98                   & 67.57          & \textbf{0.51} & 0.98                   & 64.43          & 0.33          & 0.99                   & 74.19          & 0.56          & \textbf{1}             & 66.06          & \textbf{0.46} & \textbf{1} \\
\hspace{1cm} $T5_{top-p}$     & 68.02          & 0.37          & 0.99                   & 66.27          & 0.48          & \textbf{1}             & 65.15          & 0.32          & \textbf{1}             & 71.82          & 0.54          & \textbf{1}             & 66.98          & 0.45          & \textbf{1} \\
\hspace{1cm} $BART$           & \textbf{70.71} & 0.42          & 0.94                   & 67.30          & 0.46          & 0.97                   & 68.16          & \textbf{0.33} & 0.99                   & 75.04          & 0.55          & 0.99                   & 63.95          & 0.42          & 0.92       \\
\hspace{1cm} $BART_{top-K}$   & 69.60          & \textbf{0.43} & 0.94                   & \textbf{69.47} & 0.47          & 0.96                   & \textbf{72.10} & 0.32          & 0.99                   & \textbf{76.81} & 0.56          & 0.99                   & 64.58          & 0.40          & 0.94       \\
\hspace{1cm} $BART_{top-p}$   & 67.53          & 0.40          & 0.94                   & 68.36          & 0.47          & 0.97                   & 67.35          & 0.32          & 0.99                   & 76.58          & \textbf{0.57} & 0.99                   & 65.46          & 0.41          & 0.93       \\ \bottomrule
\end{tabular}
\end{table*}

The task of translating a graph to text is predominantly a Machine Translation task. Thus, the PLM architecture of preference are seq2seq models such as Google's T5 and Facebook's BART. However, for completeness we also include an auto-regressive model - OpenAI's GPT-2 in our evaluation. The performance of these models are bench-marked across three dataset configurations: WebNLG, DART, and a combination of both. Table \ref{table:5} shows our evaluation results for these PLMs on automated word-based metrics. From this table, there are three key takeaways: For every model, the performance improves with the third dataset configuration (both the WebNLG and DART datasets). The $T5_{Base}$ model significantly outperforms the competitors while $GPT2$ falls short across all benchmarks. Finally, although $T5_{Small}$ outperforms $BART_{Large}$, their performance is almost competitive. From these observations, $T5_{Base}$ and $BART_{Large}$, with the third dataset configuration, are T\textsuperscript{3}'s preferred language models.

\subsection{\name Evaluation}
To evaluate the performance of $T^{3}$, we measure its performance with respect to our baseline - the templated generation framework. The templated generation takes in the data from Stage I of \name, however, instead of passing it to Stage II, it feeds it to a template designed for the desired domain. The narratives generated by these systems are evaluated based on three core dimensions of linguistic quality:

\begin{itemize}
    \item The Flesch's RE (Reading Ease) score \cite{re} measures the readability of a text based on the average length of its sentences and the average number of syllables of its words\footnote{\url{https://pypi.org/project/textstat/}}. Ranging from 0 to 100, increasing scores represent increasing levels of readability.
    \item The TTR (Type Token Ratio)\footnote{\url{https://pypi.org/project/lexical-diversity/}} is a measure of text diversity where the \textit{tokens} refers to the total number of words in a given text while \textit{types} refers to the number of non-repeating unique words. Simply calculated as $TTR = \frac{Types}{Tokens}$, the closer the TTR is to 1, the more lexical variety there is in a given text.
    \item The $G$ (grammar score)\footnote{\url{https://pypi.org/project/language-tool-python/}}, represents the grammatical integrity of the text. Similar to TTR, the closer $G$ is to 1, the better the grammar of the text.
    \newline $G = 1 - \frac{\textrm{Number of grammatical errors in a sentence}}{\textrm{Number of words in a sentence}}$
\end{itemize}

For each of our five datasets described in section 5-B, the RE score, TTR, and Grammar score (G) are averaged-out for the aforementioned ten countries/states. The performance of \name is evaluated with three decoding strategies: $T^3$ with $PLM_{top-K}$ represents the use of top-K sampling scheme, $T^3$ with $PLM_{top-p}$ represents the use of top-p sampling scheme, and $T^3$ with simply $PLM$ refers to the default sampling scheme where words are sampled from the base conditional probability distribution without the use of top-K or top-p strategies. Table \ref{table:6} illustrates the comparative performance of \name with templated generation. From this, we make four key observations:
\begin{enumerate}
    \item \name significantly outperforms templated generation in lexical diversity. The highest increase in lexical diversity was observed in the COVID19 dataset where \name increases the TTR by 65.38\% while the lowest observed increase was in the DOTS Exports dataset where \name increases the TTR by 13.33\%.
    \item \name remains closely competitive with templated generation in maintaining grammatical integrity. As templated generation uses pre-defined sentence planning, the grammar is expected to be perfect ($TTR=1$). While \name achieves perfect grammatical integrity in the DOTS Exports, U.S. CO Pollution, and World Population datasets, the highest observed loss in grammatical integrity was 7.9\% in the Global Temperature dataset.
    \item \name consistently outperforms templated generation in terms of readability, although not significantly. We attribute this to the distinct sentences formed when each element of the knowledge graph is translated to text.
    \item In terms of PLM selection, we observe that T5 tends to lean more towards grammatical integrity while BART tends to produce more linguistically diverse text. Similar observations are made for the sampling strategies: top-p sampling leads to more grammatical consistent texts while top-K sampling promotes linguistic diversity.
\end{enumerate}

\section{Expert Review}
We conduct an expert review $(n=21)$ \cite{expert} to validate the practicality of $T^{3}$. The review simultaneously acts as a human evaluation of $T^{3}$'s narratives as well. 85.7\% of the recruited experts had expertise in data science, 76.2\% in data visualization, and 66.7\% in NLP. When asked to rate their trust in machine-generated narratives on a 1 to 5 Likert scale, the response from the experts resembled a right-skewed bell-curve where 42.9\% of the experts had chosen a rating of 3 (neither complete trust or distrust in machine-generated narratives). In agreement with \cite{readers}, 61.9\% of the recruited experts acknowledged being dismissive of machine-generated narratives, while the remaining claimed equal treatment of both machine and human generated narratives. 

\begin{figure}[h]
    \centering
    \includegraphics[width=\linewidth]{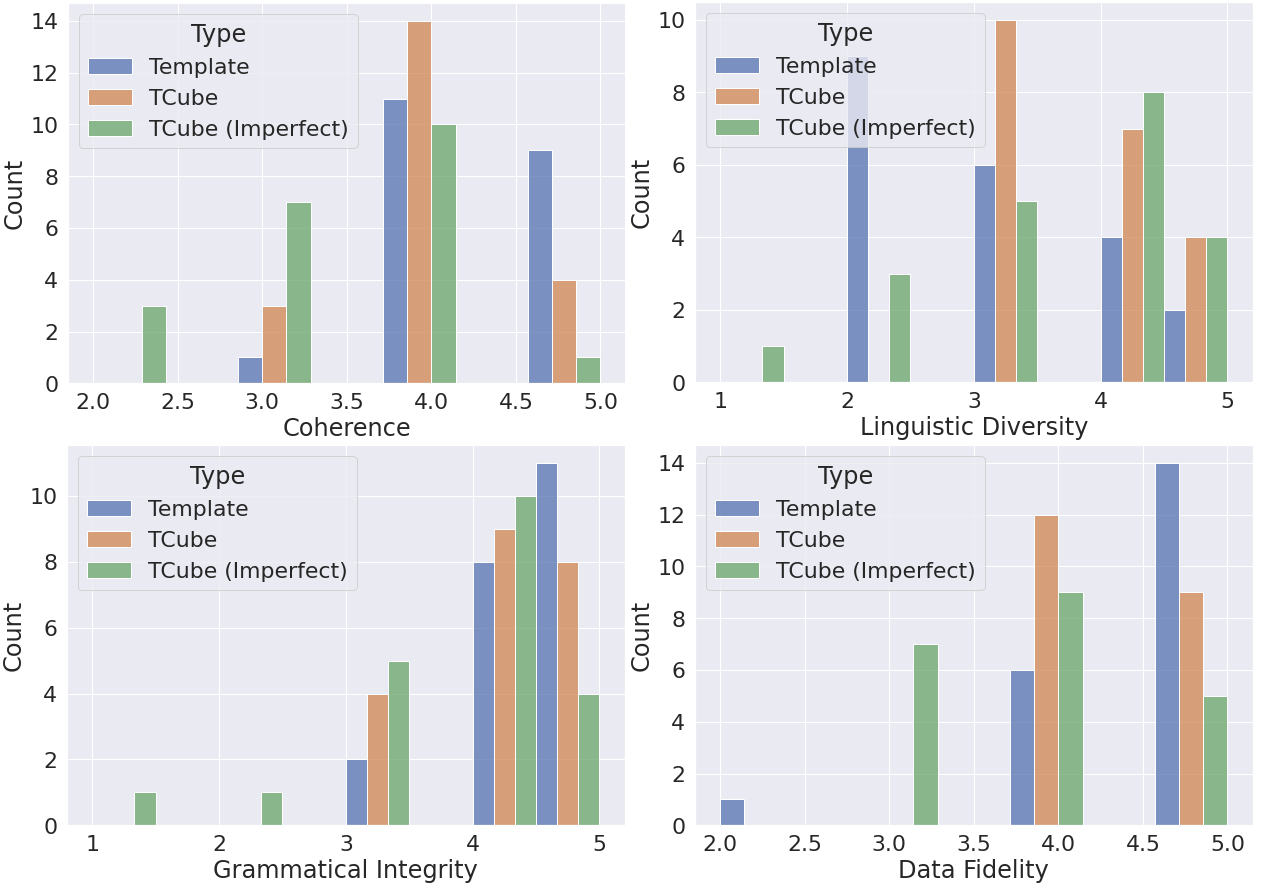}
    \caption{Histogram of Likert Ratings based on Narrative Type.}
    \label{fig:hist}
\end{figure}

The experts, each, were presented with 2 time-series datasets, where each time-series was accompanied with 4 narratives - a baseline templated narrative, 2 narratives randomly sampled from $T^{3}$, and finally, a sub-par \name narrative (generated by repeatedly sampling from \name until a a sub-par narrative was generated). For each of these narratives, the experts were asked to rate its coherence, linguistic diversity, grammatical integrity, and data fidelity (does the model tend to hallucinate?) on a 1 to 5 Likert scale. Figure \ref{fig:hist} presents an overview of the findings: $T^{3}$ and templated generation were rated comparably in terms of coherence, grammatical integrity, and data fidelity. However, $T^{3}$ was rated considerably higher in terms of linguistic diversity - in alignment with our experimental findings. In their concluding remarks, 76.2\% of the experts chose $T^{3}$ over templated narratives for deployable systems. For the remaining 23.8\% of the experts that chose templated narratives, their sentiment resonates with the need for mission-critical data fidelity.

\section{Conclusion and Future Work}
We have presented $T^{3}$, a domain-agnostic neural framework for time-series narration. Through our experiments, we outline a strategy forward for universal time-series narration. There are numerous avenues to pursue to augment the space of time-series narration. From the analysis of time-series data to the realization of natural language summaries, work in each of these space will bring us closer to better data-to-text systems. With a dataset of time-series and narrative pairs, a promising direction for future exploration lies in learning direct mappings from numbers to text, extending beyond just time-series.

\section{Acknowledgements}
This work was partially supported by DARPA (Defense Advanced Research Projects Agency) under contract number FA8650-17-C-7720. The views, opinions and/or findings expressed in this publication are solely those of the author(s).

\bibliographystyle{IEEEtran}
\bibliography{tcube}

\end{document}